\documentclass[letterpaper, 10 pt, conference]{ieeeconf}  

\IEEEoverridecommandlockouts                              

\usepackage{amsmath} 
\usepackage{amssymb}  
\usepackage{url}
\usepackage{graphicx} 
\usepackage{lipsum}    
\usepackage{algorithm}
\usepackage{algpseudocode}
\usepackage{color,soul}

\title{\LARGE \bf
Terrain-Aware Local Path Planning with Global DEM Data Integration for Autonomous UGV Navigation
}

\author{Devender Singh$^{1}$, Issah Nazif Suleiman$^{1}$, Paul Mitten$^{2}$, Glenn Cutler$^{2}$, Vinicius Prado da Fonseca$^{1}$\\ and Matthew Hamilton$^{1}$
\thanks{$^{1}$Department of Computer Science, Memorial University of Newfoundland, St. John's, NL, Canada
{\tt\small vpradodafons@mun.ca}}%
\thanks{$^{2}$Compusult Ltd., St. John's, NL, Canada}
}

\begin{document}

\maketitle
\thispagestyle{empty}
\pagestyle{empty}

\begin{abstract}
Autonomous navigation in complex outdoor terrains presents critical challenges for unmanned ground vehicles (UGVs) due to the inherent disconnect between global mapping and real-time sensor feedback.
This work proposes a hybrid framework that integrates low-resolution Digital Elevation Model (DEM) data with real-time LiDAR-based obstacle detection and terrain analysis for efficient path planning.
A global path is initially computed using a preprocessed DEM-based A* algorithm.
Subsequently, local sensor data drives adaptive path correction, enabling the UGV to negotiate sudden environmental changes while maintaining safety and efficiency.
Simulation results in Gazebo demonstrate significant improvements over a baseline approach, achieving a 95\% obstacle avoidance rate and reducing the average encountered slope from $8^\circ$ to $2.7^\circ$ in custom terrain.
This integration enhances path efficiency and terrain traversability and supports robust real-time adaptation, paving the way for more reliable autonomous navigation in dynamic outdoor environments.
\end{abstract}


\section{Introduction}

Path planning and navigation are critical in developing autonomous systems, particularly for Autonomous Unmanned Ground Vehicles (UGVs).
Improving such systems has a potential impact across various applications, including military reconnaissance~\cite{zhang2024highly}, search and rescue operations~\cite{karapetyan2024ag}, infrastructure inspection~\cite{nguyen2020practical}, and agricultural monitoring~\cite{shamshiri2018agricultural}.
However, the dynamic nature of outdoor terrains and the complexities of real-world settings demand continuous research into robust algorithms that integrate local sensor data with global map information.

Recent advancements in navigation algorithms, such as ground segmentation~\cite{ulusoy2023development, li2023mseg3d} and slope estimation~\cite{chen2023real, broggi2013terrain}, have significantly contributed to the field of autonomous navigation.
Combining these techniques can develop a more comprehensive path-planning strategy, incorporating environmental awareness to evaluate slopes, determine optimal turning angles, and assess terrain characteristics within a robot's field of view. 
This integration enhances the decision-making process in dynamic environments, improving the adaptability and intelligence of autonomous systems.

\begin{figure}[!t]
    \centering
    \includegraphics[width=\columnwidth]{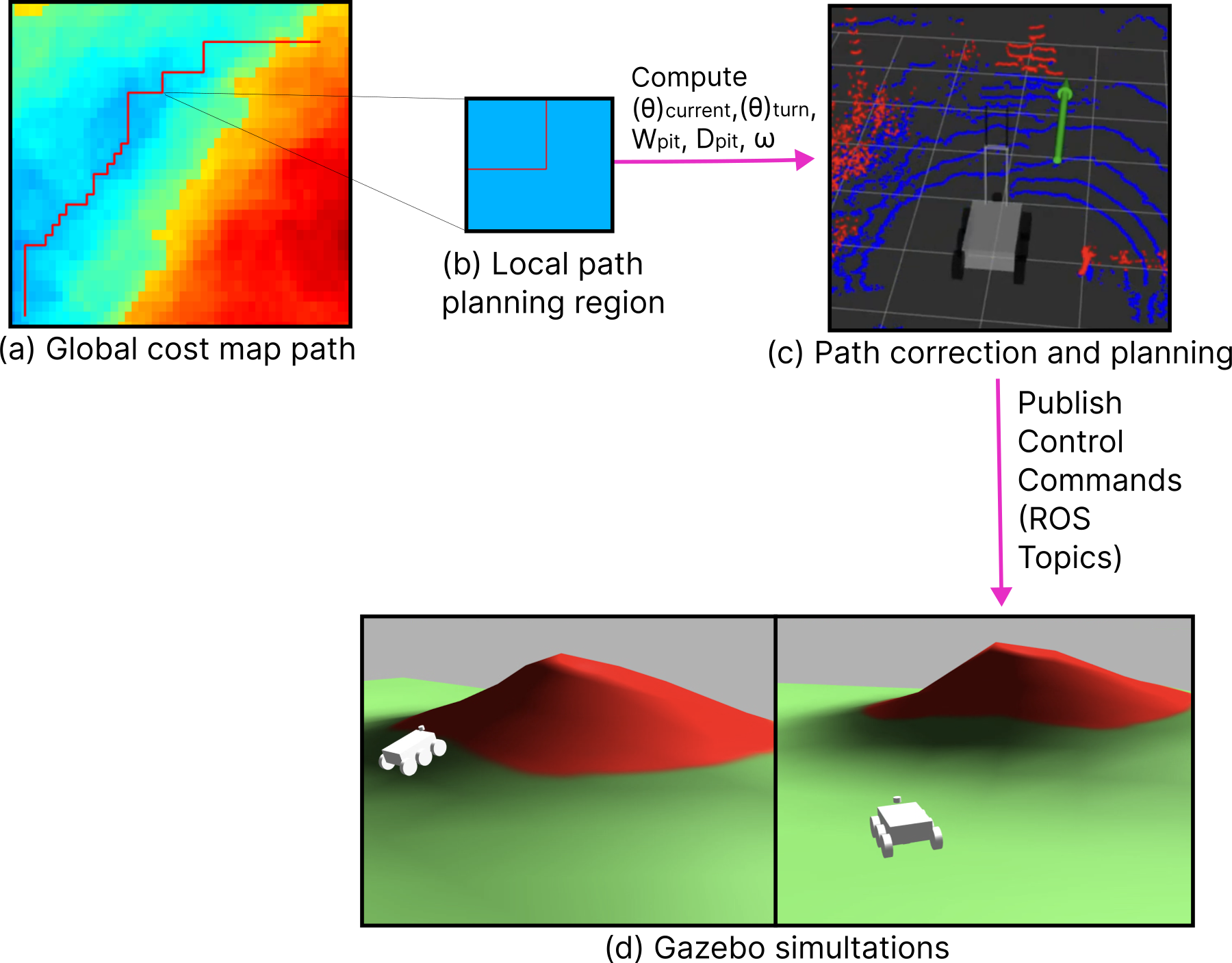}
    \caption{Local path planning with global DEM data integration. 
    (a) The global cost map path generated from SRTM-based elevation data. 
    (b) Local Path Planning Region (30m) extracted from the global map for finer trajectory 
    adjustments.
    (c) Rviz visualization of real-time LiDAR-based obstacle detection and path correction.
    (d) Gazebo simulation demonstrating the UGV avoiding obstacles in a dynamic environment (right) compared to the baseline where UGV gets stuck (left)}
    \vspace{-3mm}    
    \label{fig:figure1_final}
\end{figure}

Despite these advancements, several challenges persist. Many existing systems cannot adapt to sudden environmental changes, such as moving obstacles, in real-time~\cite{zhang2024highly}. Additionally, there remains a trade-off between safety and efficiency, with current frameworks often prioritizing one at the expense of the other. Furthermore, the scarcity of hybrid approaches that combine automated path planning  with comprehensive terrain analysis limits the effectiveness of robotic navigation in complex terrains.

\begin{figure*}[!ht]
    \centering
    \includegraphics[width=\linewidth]{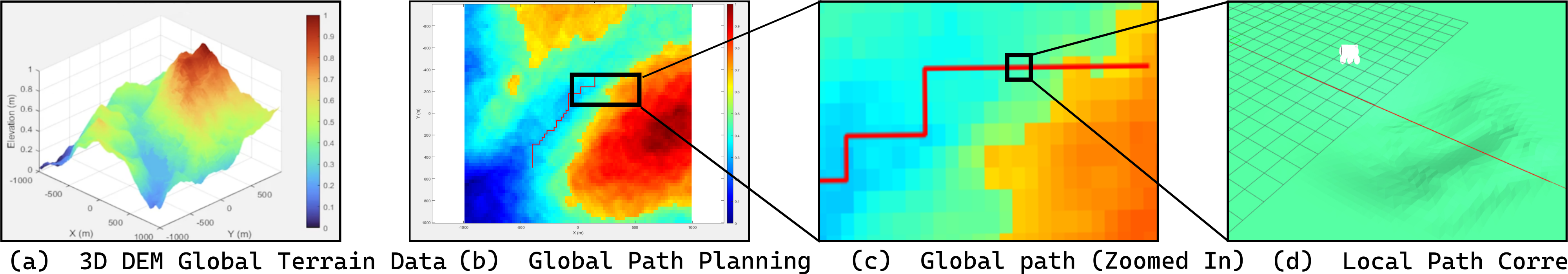}
    \caption{
    (a) Global Processing: Acquisition of DEM data, preprocessing, and generation of the global cost map; 
    (b) Global Path Planning: Optimal trajectory determination using A*; 
    (c) Zoomed-in view of the global cost map, highlighting a selected region for local path refinement; 
    (d) Local Path Correction: Integration of real-time sensor (LiDAR) data for ground segmentation, obstacle detection, and dynamic trajectory adjustment within the selected region.
    }
    \label{fig:workflow}
\end{figure*}

Fig.~\ref{fig:figure1_final} provides an overview of this study, which tackles these challenges by introducing a hybrid framework that seamlessly integrates global satellite-derived Digital Elevation Model (DEM) data with real-time sensor information.
Specifically, the present work explores two critical research questions:
(1)~How can local sensor data seamlessly integrate with global DEM maps to improve navigation through continuous cost map updates? 
(2)~How does this automated path planning compare to other navigation approaches regarding efficiency, safety, and accuracy?
By addressing these questions, this research aims to enhance theoretical frameworks for autonomous systems, bridging the gap between global planning and local execution to enable robust and efficient navigation in challenging outdoor environments.

The main aspects of our proposed pipeline are described in Fig.~\ref{fig:workflow}.
The pipeline begins with global processing (Fig.~\ref{fig:workflow}a), where DEM data is preprocessed to generate a global cost map representing large-scale terrain features.
Using this cost map, an optimal trajectory is computed via A* path planning (Fig.~\ref{fig:workflow}b).
A selected region from this trajectory is then zoomed in for further refinement (Fig.~\ref{fig:workflow}c), enabling detailed analysis at a finer resolution.
Finally, real-time sensor data is integrated to perform local path corrections through ground segmentation, obstacle detection, and dynamic trajectory adjustments (Fig.~\ref{fig:workflow}d). 

In summary, the main contributions of this work are:
\begin{itemize}
    \item We introduce a hybrid framework that integrates low-resolution DEM data with real-time local sensor information for robust UGV navigation in challenging outdoor terrains.
    \item We develop a cost map generation scheme that balances slope and elevation, augmented by penalties for extreme elevations, ensuring safer route selection.
    \item We present a local path correction mechanism using LiDAR-based ground segmentation, slope estimation, and adaptive obstacle avoidance.
    \item We validate our approach in simulation, demonstrating improved obstacle avoidance (95\% detection rate) and lower average slopes (2.7$^\circ$ vs.\ 8$^\circ$) compared to a baseline.
\end{itemize}

The next section reviews existing literature and identifies research gaps. Section~\ref{sec:methodology} presents the details of our hybrid framework, including global path planning and local corrections. Section~\ref{sec:simulation} describes the simulation environment, experimental setup, and evaluation metrics. Section~\ref{sec:results} discusses the results and compares our method to a baseline approach. Finally, Section~\ref{sec:conclusion} concludes the paper by summarizing the key findings and outlining future research directions.

\section{Related Works}

Autonomous navigation for UGVs has been a subject of extensive research, focusing on enhancing path planning, obstacle avoidance, and terrain adaptability. Significant strides have been made in global path planning, local refinement, and segmentation techniques; however, challenges remain in seamlessly integrating these components into a unified, robust framework.

Classical path-planning algorithms, such as A* and Dijkstra’s, have long been the foundation of global navigation systems due to their efficiency and optimality in structured environments \cite{9196928}. Although these methods excel in well-defined settings, they struggle in unstructured terrains with dynamic obstacles and environmental variability. More recent approaches, such as Rapidly-Exploring Random Trees (RRT) and Hybrid Potential-Based Probabilistic Roadmaps (HPPRM), offer improved flexibility for dynamic path planning but often face computational inefficiencies and heuristic reliance in large-scale environments \cite{ravankar2020hpprm}.

Although heuristics help reduce complexity, they can lead to suboptimal solutions in complex or noisy environments, especially when real-time sensor data and environmental nuances are not adequately captured \cite{wilt2016effective}. Moreover, these methods typically require extensive parameter tuning and domain expertise, limiting their robustness in novel scenarios. Despite these advancements, there is limited focus on integrating real-time local sensor data to refine paths generated from global maps.


Obstacle avoidance has been extensively studied, with LiDAR-based segmentation \cite{gomes2023survey} and visual-based detection methods such as YOLO \cite{redmon2016you} being widely adopted. These methods demonstrate high accuracy in identifying static and dynamic obstacles but often fail to adapt to terrain-specific challenges such as negative obstacles, including pits and ditches. Recent work on traversability estimation, such as the G-VOM system, combines voxel-based mapping with obstacle detection and slope analysis, enabling off-road navigation \cite{overbye2022g}. However, these approaches frequently lack integration with global path-planning strategies, leaving room for improvement in balancing local adaptability with global efficiency.

Slope estimation is another crucial aspect of autonomous navigation, particularly in off-road conditions. Traditional methods typically rely on gradient-based calculations over static grids, which often fail to capture the complexities of rugged and sloped terrains \cite{broggi2013terrain}. Advanced approaches employing plane fitting and historical slope data analysis \cite{shetty2022road} provide better accuracy, but their real-time applicability in dynamic environments remains limited. Furthermore, the absence of adaptive thresholding mechanisms in these systems restricts their ability to generalize across varying terrains.

Local path refinement techniques, such as Simultaneous Localization and Mapping (SLAM) with cost map updates, have shown promise for dynamically adjusting paths based on real-time sensor data \cite{bresson2017simultaneous}. However, most SLAM-based methods prioritize local consistency at the expense of maintaining global path alignment, resulting in suboptimal navigation outcomes in larger, complex environments. This gap highlights the need for a hybrid approach that combines global optimization with local refinement.

Despite significant advancements in individual aspects of navigation, few studies provide a unified framework that integrates global path optimization, real-time segmentation, obstacle avoidance, slope estimation, and dynamic local refinement. As shown in Table~\ref{tab:comparison}, our approach bridges these gaps by seamlessly combining global and local components, enabling robust navigation across diverse terrains and dynamic conditions.

In summary, while existing approaches have made substantial contributions to individual aspects of autonomous navigation, the integration of these components into a cohesive system remains underexplored. Our work addresses this limitation by introducing a framework that combines the strengths of global optimization, real-time terrain analysis, and adaptive local path refinement to enable efficient and safe navigation in complex environments.

\begin{table}[h!]
\centering
\caption{Comparison of Our Approach with Related Works}
\label{tab:comparison}
\begin{tabular}{|l|c|c|c|c|c|}
\hline
\textbf{Reference} & \textbf{OA} & \textbf{SE} & \textbf{RTS} & \textbf{GPO} & \textbf{LPR} \\
\hline
\cite{zhang2024highly}  & \checkmark &  &  &  &  \\
\cite{overbye2022g}  & \checkmark & \checkmark &  &  &  \\
\cite{broggi2013terrain}  &  & \checkmark &  &  &  \\
\cite{shetty2022road}  &  & \checkmark &  &  &  \\
\cite{ulusoy2023development}  &\checkmark  &  & \checkmark &  &  \\
\cite{li2023mseg3d}  &  &  & \checkmark &  &  \\
\cite{9196928}  &  &  &  & \checkmark & \checkmark\\
\cite{ravankar2020hpprm}  &  &  &  & \checkmark &  \\

\hline
\textbf{Our Approach} & \checkmark & \checkmark & \checkmark & \checkmark & \checkmark \\
\hline
\end{tabular}
\end{table}

\section{Methodology}\label{sec:methodology}

This section describes our hybrid framework for UGV navigation that integrates global path planning using low-resolution DEM data with real-time local path corrections. The methodology comprises three main components: global path planning, local path correction and planning, and dynamic adjustments for integration of global and local paths.


\subsection{Global Path Planning}

The global path planning module uses preprocessed Shuttle Radar Topography Mission (SRTM) 1 Arc-Second Global DEM data (GeoTIFF format from USGS Earth Explorer) to generate an initial trajectory over a large-scale terrain centered on St. John’s, Newfoundland and Labrador, Canada. While SRTM’s 30\,m resolution provides broad terrain trends, it cannot capture fine-grained obstacles (e.g., rocks, ditches, pits) or rapid elevation changes—a limitation addressed through our local correction framework. In this study, the area of interest (AOI) is defined as a circular region with radius \( r_m = 1000\,\mathrm{m} \). The processing pipeline consists of the following steps:

\subsubsection{DEM Preprocessing}\label{subsubsec:dem}

To define the region of interest (ROI), we use the center coordinates $\phi_c = 47.4870^\circ$ (latitude) and $\lambda_c = -52.8629^\circ$ (longitude). The metric conversion factors for latitude and longitude are given by $m_\phi = 111320$ m/deg and $m_\lambda = 111320\cos\phi_c$ m/deg. Using these, the angular extents of the ROI are computed as $\Delta\phi = r_m / m_\phi$ and $\Delta\lambda = r_m / m_\lambda$. 

From these values, the latitude and longitude limits are determined as $\phi_{\min} = \phi_c - \Delta\phi$, $\phi_{\max} = \phi_c + \Delta\phi$, $\lambda_{\min} = \lambda_c - \Delta\lambda$, and $\lambda_{\max} = \lambda_c + \Delta\lambda$. The DEM data within these bounds is cropped, interpolated to a finer resolution, and normalized to produce a 3D surface representation, $z(x,y)$, which serves as the basis for subsequent cost calculations.

\subsubsection{Cost Map Generation}\label{subsubsec:costmap}

Using the normalized elevation function $z(x,y)$, the local slope magnitude is computed as
\begin{align}
s(x,y) = \sqrt{\left(\frac{\partial z(x,y)}{\partial x}\right)^2 + \left(\frac{\partial z(x,y)}{\partial y}\right)^2}.
\end{align}
The cost function is defined by
\begin{align}
C(x,y) = w_s\, s(x,y) + w_z\, z(x,y),
\end{align}
where the weights $w_s = 1.0$ and $w_z = 0.5$ have been chosen empirically to balance the influence of slope and elevation. An adaptive penalty \( P = 10 \) is applied to elevation extremes to incorporate a slight cost bias against extreme low-lying areas to avoid potential flood-prone or unsafe regions. This ensures the UGV avoids areas where elevation extremes may indirectly compromise safety, even if slopes are manageable~\cite{guastella2017global}. Finally \( C(x,y) \) is normalized to \( [0,1] \) standardizing cost aggregation for path planning step. The 30\,m DEM resolution inherently limits cost map fidelity, as seen in Figure~\ref{fig:workflow}(c), where elevation thresholds (\( z_H \), \( z_L \)) smooth over sub-resolution terrain variations. This necessitates local sensor-driven refinements during navigation.

\subsubsection{Path Generation}\label{subsubsec:path_generation}

The A* algorithm computes the optimal path by minimizing cumulative costs derived from the DEM. The trajectory prioritizes safety-critical factors encoded in the cost map, such as steep slopes and elevation-linked hazards, ensuring adherence to the UGV’s mechanical and operational limits. 

\subsection{Local Path Correction and Planning}

To refine the global path for small-scale, uncertain terrains, a local correction module is employed. This module uses real-time sensor data to dynamically adapt the path to ensure stable traversal across varying surface conditions. The process involves:

\begin{enumerate}
    \item \textbf{Real-time LiDAR Data Processing}: Segmentation of terrain features using Ground Plane Fitting (GPF) and RANSAC.
    \item \textbf{Slope Estimation and Adaptive Traversability Analysis}: Evaluation of terrain slopes with dynamic thresholding.
    \item \textbf{Obstacle Avoidance and Adaptive Steering}: Detection of pits, bumps, and obstacles with turning decisions guided by distribution-based analysis.
\end{enumerate}

This approach ensures that the UGV navigates diverse terrains effectively, avoiding hazards while maintaining an optimized local path.

\subsubsection{Real-Time LiDAR Data Processing}
Real-time LiDAR point cloud data is segmented into ground and non-ground points using a hybrid Ground Plane Fitting (GPF) approach combined with iterative RANSAC. The process is as follows:

\begin{itemize}
    \item \textbf{Initial Ground Seed Extraction:}  
    The lowest $N_{\mathrm{LPR}} = 250$ points in the cloud (sorted by elevation) form the low point reference (LPR). Points within $\tau_{\mathrm{seed}} = 0.8$~m above the LPR are selected as initial ground seeds.
    \item \textbf{Iterative Plane Fitting:}   A plane is fitted to the current ground seeds using RANSAC. In each iteration, the estimated plane is defined as
\begin{align}
    n \cdot x + d = 0 
\end{align}
    where $n$ is the estimated normal vector and $d$ is the offset. The absolute distance from every point in the cloud to the plane is calculated. Points with a distance less than the threshold $\tau_{\mathrm{dist}} = 0.1\,\mathrm{m}$ are classified as ground. These points form a refined seed set for the next iteration. This iterative procedure is typically repeated for 3 iterations to improve robustness.
    \item \textbf{Ground and Non-Ground Classification:}  
    After refinement, points consistently within $\tau_{\mathrm{dist}}$ of the plane are classified as ground, while the rest are labeled as non-ground (obstacles). These thresholds were empirically selected for robustness against noise and terrain variations.
\end{itemize}

\subsubsection{Slope Estimation and Adaptive Traversability Analysis}
\begin{figure}[!t]
    \centering
    \includegraphics[width=\columnwidth]{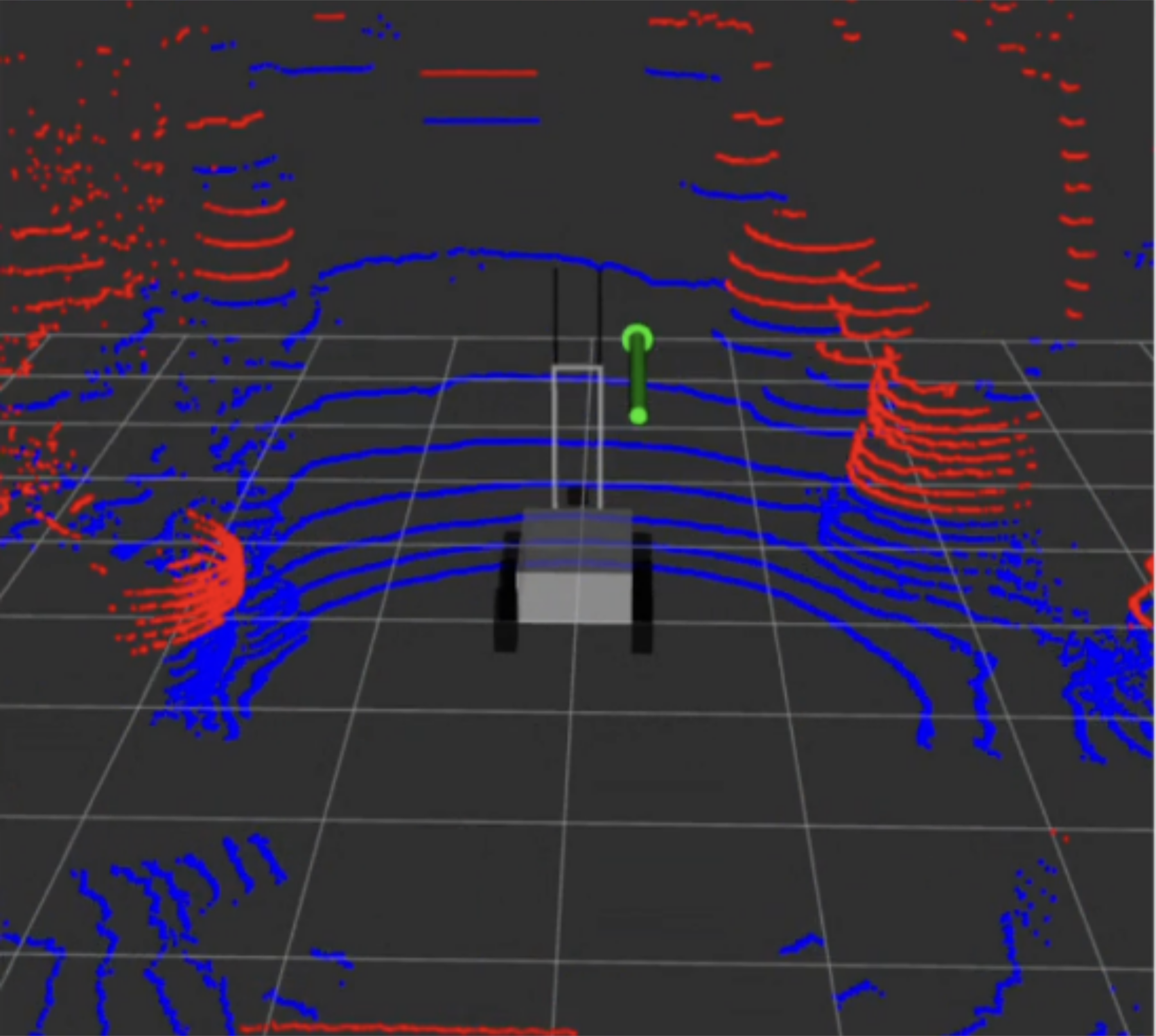}
    \caption{Visualization of ground (blue points) and non-ground (red points) segmentation from LiDAR data, along with the final path direction (green marker) as visualized in RViz. This image illustrates the result of our local path correction approach, where the UGV dynamically adjusts its trajectory to avoid nearby obstacles and maintain safe navigation through complex terrains.}
    \label{fig:segmentation_path}
\end{figure}

The UGV estimates the current slope, denoted as $\theta_{\text{current}}$, using the ground plane normal $\vec{n}$ and the vertical axis $\vec{z}=[0,0,1]^T$:
\begin{equation}
    \theta_{\text{current}} = \arccos\!\Bigl(\frac{\vec{n}\cdot \vec{z}}{\|\vec{n}\|\|\vec{z}\|}\Bigr).
\end{equation}
We maintain a sliding window $s$ of the most recent $N=40$ slope measurements to capture typical variations in terrain over time. After inserting $\theta_{\text{current}}$ into $s$ (and removing the oldest slope if $|s|>N$), we compute the mean and standard deviation:
\begin{align}
    \mu_\theta = \mathrm{mean}(s), 
    \quad 
    \sigma_\theta = \mathrm{std}(s).
\end{align}

\noindent
\textbf{Dynamic Thresholds.}
We define two slope thresholds based on the distribution of recent slopes:
\begin{equation}
    \tau_{\mathrm{warn}} = \mu_\theta + \sigma_\theta, 
    \quad 
    \tau_{\mathrm{crit}} = \mu_\theta + 3\,\sigma_\theta.
\end{equation}
\begin{itemize}
    \item \(\tau_{\mathrm{warn}}\) flags when $\theta_{\text{current}}$ exceeds one standard deviation above the mean, indicating steeper terrain than usual.
    \item \(\tau_{\mathrm{crit}}\) (three standard deviations above the mean) covers roughly the top $0.15\%$ of expected slopes in a normal distribution, signaling extremely steep terrain.
\end{itemize}

\noindent

Algorithm~\ref{alg:slope} summarizes this procedure, showing how slope history is tracked and how the UGV’s response escalates with increasingly steep terrain.
\begin{algorithm}[!ht]

\begin{algorithmic}[1]
\Require Ground plane normal $\vec{n}$, vertical axis $\vec{z}=[0,0,1]^T$, sliding window of slopes $s$ (size $N=40$)
\Statex
\State \textbf{Compute current slope:} 
\[
\theta_{\text{current}} \gets \arccos\!\Bigl(\tfrac{\vec{n}\cdot \vec{z}}{\|\vec{n}\|\|\vec{z}\|}\Bigr)
\]
\State Append $\theta_{\text{current}}$ to $s$. If $|s| > N$, remove the oldest measurement.
\State \textbf{Compute mean and std:} 
\[
\mu_\theta \gets \mathrm{mean}(s), 
\quad
\sigma_\theta \gets \mathrm{std}(s)
\]
\State \textbf{Set dynamic thresholds:} 
\[
\tau_{\mathrm{warn}} \gets \mu_\theta + \sigma_\theta, 
\quad 
\tau_{\mathrm{crit}} \gets \mu_\theta + 3\,\sigma_\theta
\]
\If{$\theta_{\text{current}} > \tau_{\mathrm{crit}}$}
    \State \textbf{Emergency maneuver} (halt + high-turn escape or halt then back up)
\ElsIf{$\theta_{\text{current}} > \tau_{\mathrm{warn}}$}
    \State \textbf{Obstacle avoidance} (Section III.B.3)
\Else
    \State \textbf{Continue nominal motion}
\EndIf
\end{algorithmic}
\caption{Slope Estimation and Adaptive Traversability Analysis}\label{alg:slope}
\end{algorithm}

\subsubsection{Obstacle Avoidance and Adaptive Steering}\label{subsubsec:obstacle_avoidance}
To robustly avoid obstacles, the UGV leverages non‐ground points (obtained from LiDAR segmentation) to model terrain irregularities—such as pits and bumps—as bump distributions. In our UGV coordinate frame, let $x$ denote the forward distance and $y$ the lateral displacement. The obstacle width is defined as: 
\begin{equation}
\begin{aligned}
     W_{\mathrm{pit}} = y_{\max} - y_{\min},
\end{aligned}
\end{equation}

where $y_{\max}$ and $y_{\min}$ are the maximum and minimum lateral positions of the detected non‐ground points. The closest approach is quantified by: 
\begin{equation}
\begin{aligned}
D_{\mathrm{pit}} = \min\{x\}, 
\end{aligned}
\end{equation}
with $x$ measured in meters.

We model the height profile of the obstacle as a probability density function (PDF) $h(x)$, where $h(x)$ represents the height (positive for bumps, negative for pits) at a given forward position $x$. The statistical properties of $h(x)$—its mean $\mu_h$, median, and standard deviation $\sigma_h$—are computed from the detected points. Pearson's skewness coefficient is then defined as:
\begin{align}
S = \frac{3(\mu_h - \text{median}(h))}{\sigma_h}.
\end{align}
An empirically determined threshold $\tau=0.2$ is used to classify the distribution: if $|S| \ge \tau$, the distribution is considered skewed; otherwise, it is assumed to be symmetric (normal).

\textbf{Turning Decision:}  
\begin{itemize}
    \item \textbf{Case 1: Skewed Distribution} ($|S| \ge \tau$):  
    \[
    \text{Turn} =
    \begin{cases}
        \text{Left}, & \text{if } S > 0,\\[1ex]
        \text{Right}, & \text{if } S < 0.
    \end{cases}
    \]
    \item \textbf{Case 2: Normal Distribution} ($|S| < \tau$):  
    The obstacle is sampled on both sides. Let $h_L$ and $h_R$ denote the median heights of the obstacle points on the left and right sides, respectively. Then,
    \[
    \text{Turn} =
    \begin{cases}
        \text{Right}, & \text{if } h_L > h_R,\\[1ex]
        \text{Left}, & \text{if } h_L < h_R.
    \end{cases}
    \]
\end{itemize}
\begin{figure}[!t]
    \centering
    \includegraphics[width=0.95\linewidth, height=0.45\textheight, keepaspectratio]{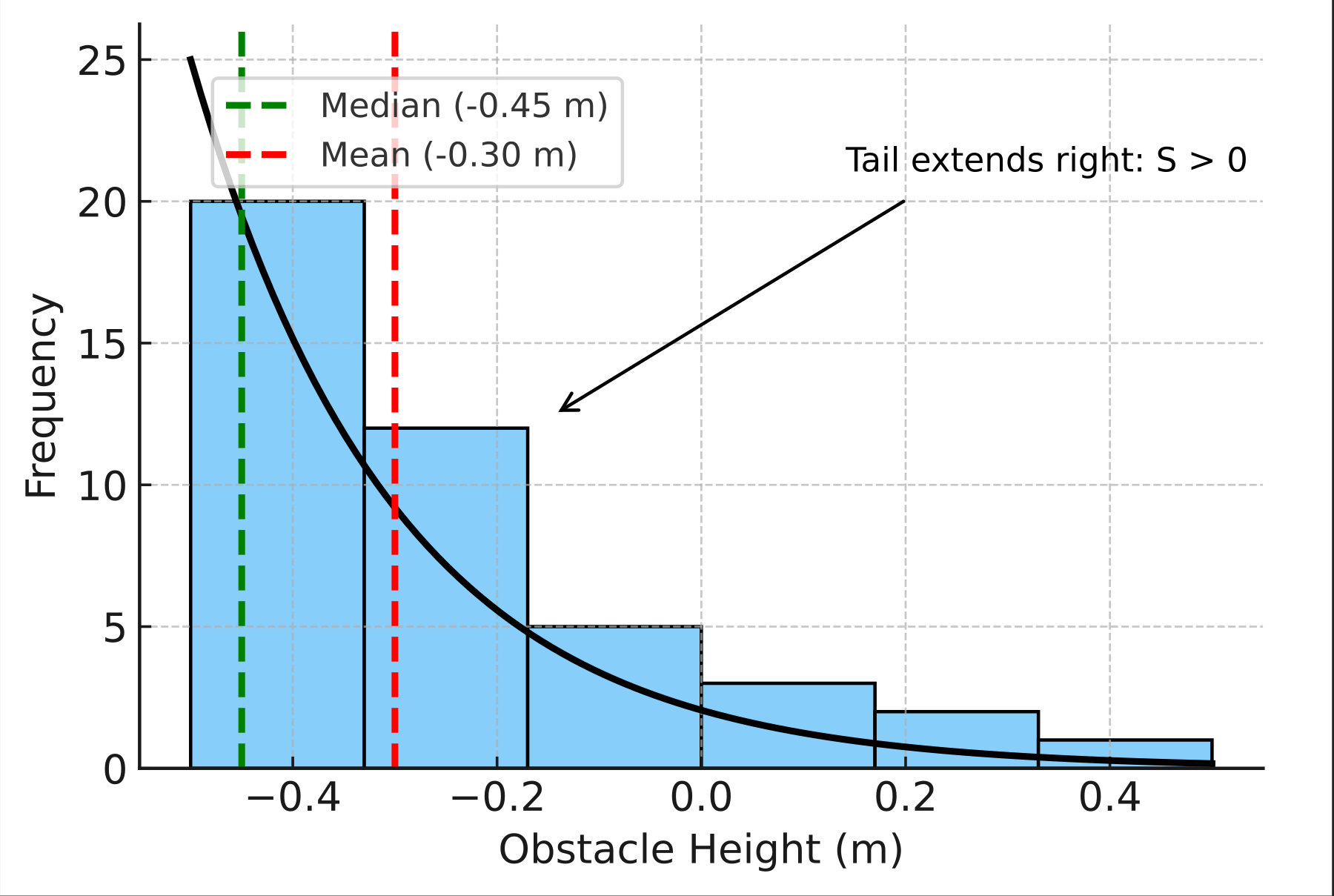}
    \caption{Height distribution from terrain in Figure 4(c) as observed by the UGV. The blue histogram shows that most obstacle heights cluster near –0.50\,m, with a right tail extending into positive values. The green dashed line indicates the median (–0.45\,m) and the red dashed line shows the mean (–0.30\,m), confirming a right-skewed distribution (\(S>0\)).}
    \label{fig:skew_terrainC}
\end{figure}

\textbf{Trajectory Correction and Command Adjustment:}  
After the turning direction is determined, the UGV computes the required turning angle as
\begin{equation}
    \theta_{\text{turn}} = \arctan\left(\frac{W_{\mathrm{pit}}}{D_{\mathrm{pit}}}\right)
\end{equation}

The angular velocity command is then given by
\begin{equation}
    \omega = \text{sgn}(S) \cdot \min\Bigl(\omega_{\max},\, \omega_{\text{base}} \cdot \theta_{\text{turn}}\Bigr),
\end{equation}

where $\omega_{\text{base}}$ and $\omega_{\max}$ denote the nominal and maximum turn rates, respectively. Concurrently, the UGV dynamically adjusts its linear velocity to maintain stability. If $D_{\mathrm{pit}}$ falls below a critical threshold (e.g., $0.5\,\mathrm{m}$), an emergency escape maneuver is triggered:
\begin{equation}
v = v_{\max}, \quad \omega = 5\,\omega_{\text{base}}.
\end{equation}

This framework enables the UGV to make rigorous, data-driven decisions for obstacle avoidance and adaptive steering in complex, dynamic terrains. Figure~\ref{fig:segmentation_path} illustrates the result of our local path correction module. The visualization shows segmented ground points (blue), non-ground points (red), and the final path direction (green marker) as determined by integrating real-time LiDAR data processing, slope estimation, and obstacle avoidance. 

\subsection{Dynamic Path Adjustment}
While the local path planning module ensures obstacle avoidance and traversability, the UGV must also maintain alignment with its desired trajectory. Dynamic path adjustment is implemented as a corrective control mechanism that continuously refines the UGV’s movement, ensuring it returns to the intended path after deviations.

The system monitors the vehicle’s lateral displacement relative to the desired trajectory (assumed to be along $y = 0$) and applies proportional angular corrections based on the computed error. This correction is governed by:

\begin{equation}
    \omega_{\text{corr}} = k_p \cdot \tan^{-1}(y_{\text{error}})
\end{equation}

where $\omega_{\text{corr}}$ is the angular correction, $k_p$ is a proportional gain, and $y_{\text{error}}$ represents the lateral deviation from the target path.

To prevent excessive oscillations, the angular correction is constrained within a maximum bound:

\begin{equation}
    \omega = \max(-\omega_{\text{max}}, \min(\omega_{\text{max}}, \omega_{\text{corr}}))
\end{equation}

The UGV continues moving forward at a predefined linear velocity while applying steering corrections. If the lateral deviation is within a predefined threshold, no correction is applied. This ensures that the vehicle follows the smoothest possible trajectory with minimal deviations.

Additionally, an override mechanism is incorporated to allow external modules to disable path correction when higher-priority navigation commands, such as obstacle avoidance, take precedence. This prevents conflicts between multiple navigation strategies and maintains a balance between path adherence and obstacle responsiveness.


\begin{figure*}[!t]
    \centering
    \includegraphics[width=\textwidth]{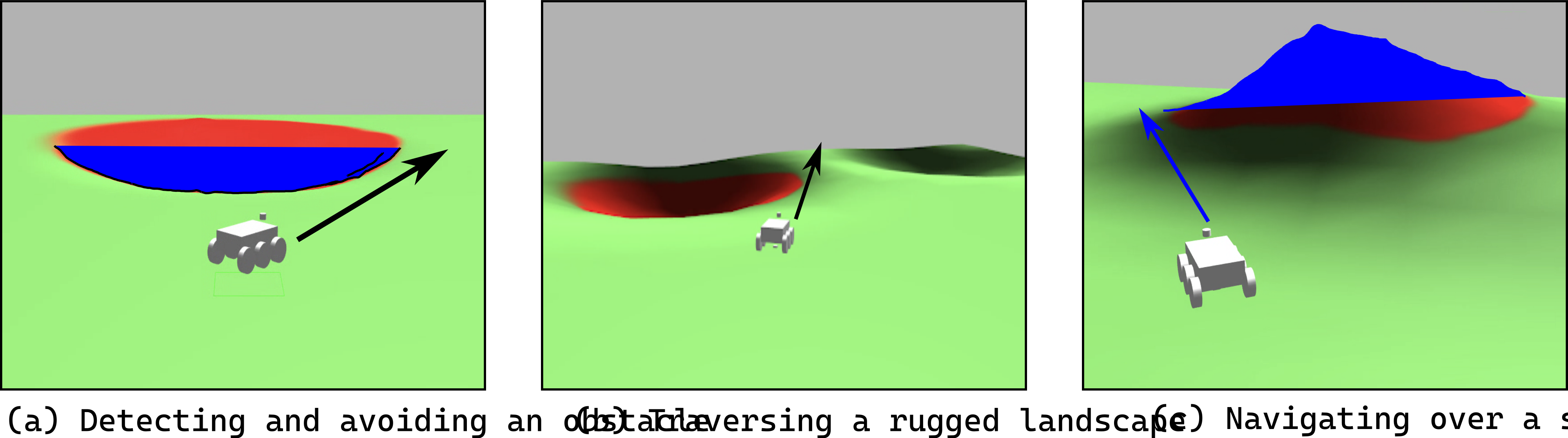}
    \caption{
    (a) UGV navigating a flat terrain while detecting and avoiding an obstacle (here a pit). The trajectory taken by the UGV is indicated with the blue arrow avoiding the obstacle in its view. 
    (b) UGV traversing a rugged landscape with visible obstacles (pits). The illustration highlights adaptive path planning and obstacle avoidance using our algorithm in action on rough terrain. 
    (c) UGV navigating over a slope, illustrating its capability to handle inclines. The figure depicts how using our algorithm, UGV detects the bump and the adjustments made to avoid and maintain stability.
}
    \label{fig:slope}
\end{figure*}
\section{Simulation and Experimental Setup}\label{sec:simulation}

\subsection{Simulation Environment}

\begin{figure}[!ht]
    \centering
    \includegraphics[width=\columnwidth, height=0.3\textheight, keepaspectratio]{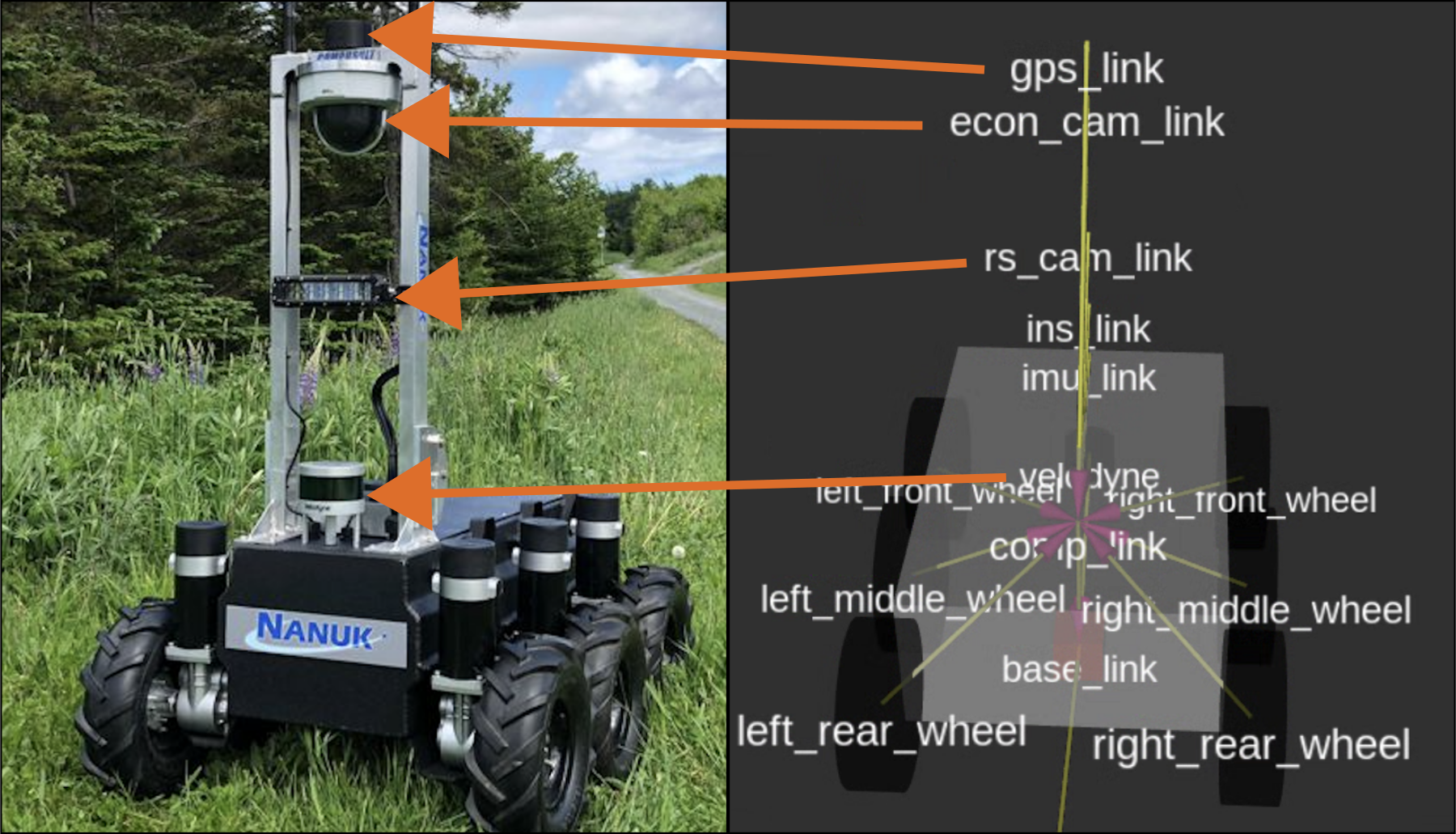}
    \caption{Side-by-side comparison of the real Compusult Ltd.’s Nanuk UGV (left) and its ROS model derived from the URDF file (right). This comparison demonstrates the fidelity of our simulation model for sensor integration and navigation evaluation.}
    \label{fig:ugv_comparison}
\end{figure}

Our experiments were conducted in a simulated environment generated using Blender. The terrains were either procedurally generated or handcrafted to include a wide variety of challenging features, such as deep pits, steep slopes, and non-rough, uneven surfaces. Specific test scenarios included:
\begin{itemize}
    \item \textbf{Pits and Negative Obstacles}: Terrains with abrupt depressions and holes to evaluate the UGV's ability to detect and avoid pitfalls.
    \item \textbf{Steep Slopes and Hills}: Gradual and extreme inclines were incorporated to test slope estimation, traversability analysis, and emergency maneuvers.
    \item \textbf{Non-Rough Terrain}: Planar regions with no obstacles.
\end{itemize}
\subsection{Evaluation Metrics}
We evaluate our approach using three key metrics:

\subsubsection{Path Efficiency}
\begin{itemize}
    \item \textbf{Total Path Length}: Measures the distance traveled relative to the optimal global path, where longer paths imply detours for obstacle avoidance.
    \item \textbf{Execution Time}: Captures the total navigation duration, highlighting the balance between safety and efficiency.
\end{itemize}

\subsubsection{Safety}
\begin{itemize}
    \item \textbf{Obstacle Clearance}: The minimum distance from obstacles during navigation, with higher clearances indicating safer paths.
    \item \textbf{Slope Traversability}: Assessed by the average slope encountered; lower slopes suggest better hazard avoidance.
\end{itemize}

\subsubsection{Accuracy}
\begin{itemize}
    \item \textbf{Path Deviation}: The lateral error from the planned global trajectory, where smaller deviations reflect more accurate navigation.
    \item \textbf{Success Rate}: The fraction of trials in which the UGV reaches its destination without collisions.
\end{itemize}


\subsection{Implementation Details}
Our implementation is built upon the Robot Operating System (ROS) framework. The system comprises two primary modules:
\begin{itemize}
    \item \textbf{Local Path Correction and Obstacle Avoidance:}  
    A ROS node processes real-time LiDAR data to segment ground from non-ground points using a hybrid Ground Plane Fitting (GPF) with iterative RANSAC. It estimates terrain slopes, detects hazards (e.g., pits, steep slopes), and computes turning commands. Emergency maneuvers are triggered when slopes exceed critical thresholds.
\end{itemize}
\begin{itemize}
    \item \textbf{Dynamic Path Adjustment:}  
    A separate ROS node monitors the UGV’s odometry to compute lateral deviation from the desired trajectory (assumed along $y=0$) and applies a proportional controll steering.An override signal is used to disable path correction during active obstacle avoidance.
\end{itemize}

The system was validated in Gazebo, where the UGV model was spawned with Blender-generated terrains. Sensor integration, including LiDAR for local perception and odometry for state estimation, was managed via ROS topics. Custom Python scripts enabled safe navigation while following the global path. The simulation includes Compusult Ltd.'s Nanuk UGV URDF for realism. Figure~\ref{fig:ugv_comparison} compares the actual Nanuk UGV with its ROS model.





\section{Results and Discussion}\label{sec:results}

\subsection{Performance Metrics}
We evaluate our approach using efficiency, safety, and accuracy metrics. Experiments were conducted on two terrain types: (i) non-rough, obstacle-free terrain and (ii) challenging terrain with steep slopes and pits. Table~\ref{tab:results} summarizes the performance of our method versus a baseline straight trajectory.

\begin{table}[ht!]
    \caption{Performance comparison of our approach versus the baseline under varying terrain conditions. NB:* --- Not applicable}
    \label{tab:results}
    \centering
    \resizebox{\columnwidth}{!}{ 
    \begin{tabular}{|l|cc|cc|}
        \hline
        \textbf{Metric} & \multicolumn{2}{c|}{\textbf{Our Approach}} & \multicolumn{2}{c|}{\textbf{Baseline}} \\
        \cline{2-5}
                        & Non-Rough & Obstacle & Non-Rough & Obstacle \\
        \hline
        Avoidance Rate (\%)    & *    & 95    & *    & 0 \\
        Average Slope (°)       & 0.7  & 2.7   & 0.7  & 8 \\
        Time to Goal (s)        & 43   & 77    & 43   & Stuck \\
        Path Deviation (m)      & 0    & 4.25  & 0    & Stuck \\
        \hline
    \end{tabular}}
\end{table}

\subsection{Comparison with Baseline}
Our approach offers notable advantages over the baseline:

\begin{itemize}
    \item \textbf{Obstacle Avoidance Rate (\%):} Our approach achieves an obstacle avoidance rate of 95\% in challenging (obstacle) scenarios, effectively detecting and avoiding pits and steep slopes. In contrast, the baseline approach shows an avoidance rate of 0\% under the same conditions. For non-rough terrain, this metric is not applicable (denoted as “*”) for both methods.
    
    \item \textbf{Average Slope (°):} On non-rough terrain, both methods maintain an average slope of 0.7°. However, in the presence of obstacles, our method reduces the average encountered slope to 2.7° compared to 8° for the baseline, indicating better management of steep terrain.
    
    \item \textbf{Time to Goal (s):} Both approaches take 43 seconds to reach the goal on non-rough terrain. Under challenging conditions, our approach requires 77 seconds—an acceptable increase given the environmental complexity, whereas the baseline becomes stuck.
    
    \item \textbf{Path Deviation (m):} On non-rough terrain, both methods exhibit zero deviation. In obstacle scenarios, our approach shows a 4.25 m deviation as it maneuvers around hazards, while the baseline again becomes \emph{stuck}.
\end{itemize}

These results show that while the baseline performs well in simple environments, it fails in challenging conditions. In contrast, our approach enhances safety and robustness by detecting obstacles, avoiding hazards, and maintaining a lower slope, despite increased execution time and path deviation.



\section{Conclusion And Future Work}\label{sec:conclusion}

This study presents a hybrid UGV navigation framework that integrates global satellite-derived DEM data with real-time local sensor information. It combines global terrain mapping with dynamic local path correction to enhance navigation in challenging outdoor environments. The system initially plans a trajectory using low-resolution DEM data, avoiding high-elevation regions, and then refines the path using local sensors to detect rough terrain, obstacles, and pits.

Results show significant improvements in navigation efficiency, safety, and accuracy, with a 95\% pit detection rate (vs. 0\% baseline) and a reduction in average slope from 8° to 2.7°. While the approach leads to a higher path deviation (4.25 m vs. 0 m baseline), this trade-off enhances obstacle detection and safe traversal.

Future work includes field trials on Compusult Ltd.'s Nanuk UGV models to validate real-world performance, integrating multi-resolution DEM data for improved planning, and exploring reinforcement learning for adaptive decision-making. These advancements aim to bridge the gap between simulation and real-world deployment.

\bibliographystyle{IEEEtran}
\bibliography{biblio}

\end{document}